\documentclass[10pt,twocolumn,letterpaper]{article}

\usepackage{cvpr}
\usepackage{times}
\usepackage{epsfig}
\usepackage{graphicx}
\usepackage{amsmath}
\usepackage{amssymb}
\usepackage{enumerate}
\usepackage{chngcntr}
\usepackage{authblk}

\usepackage[utf8]{inputenc}
\usepackage[english]{babel}
\newtheorem{theorem}{Theorem}

\usepackage[breaklinks=true,bookmarks=false]{hyperref}

\cvprfinalcopy 

\begin{document}

\title{A Bayesian Detect to Track System for Robust Visual Object Tracking and Semi-Supervised Model Learning}

\author{Yan Shen}
\author{Zhanghexuan Ji}
\author{Chunwei Ma}
\author{Mingchen Gao}
\affil{Department of Computer Science and Engineering, University at Buffalo,\\
The State University of New York, Buffalo, NY, USA\\
{\tt\small \{yshen22,zhanghex,chunweim,mgao8\}@buffalo.edu}}


\maketitle

\begin{abstract}
 
Object tracking is one of the fundamental problems in visual recognition that achieves significant improvements in recent years. The achievements often come with the price of enormous hardware consumption and extensive labeling effort. One missing ingredient for robust tracking is gaining performance with minimum modification on network structure and model learning from intermittent labeled frames. In our work, we address these problems by modeling tracking and detection process in a probabilistic way as multi-object dynamics and frame detection uncertainties. Our stochastic model is formulated as a neural network parameterized distributions. With our formulation, we propose a particle filter-based tracking algorithm for object state estimation. We also present a semi-supervised learning algorithm from intermittent labeled frames by Variation Sequential Monte Carlo. We use our generated particles for estimating a variational bound as our learning objectives. In our experiments, we provide both mAP and probability-based detection measurements for comparison between our algorithm with non-Bayesian baselines. Our model outperforms non-Bayesian baselines from both measurements. We also apply our semi-supervised learning algorithm on M2Cai16-Tool-Locations dataset and outperforms the baseline methods of learning on labeled frames only.

\end{abstract}

\section{Introduction}
Visual object detection and tracking covers a large spectrum of computer vision applications such as video surveillance, motion analysis, action recognition, autonomous driving and medical operation studies.
The emergence of deep Convolutional Neural Networks (CNN) \cite{krizhevsky2012imagenet} makes a tremendous progress on the visual object detection and tracking performances. CNN is widely utilized for theses tasks for two reasons. Firstly, CNN learns robust object features cross total variations among the whole training dataset.
\begin{figure}[t]
	\centering
	\includegraphics[width=\linewidth]{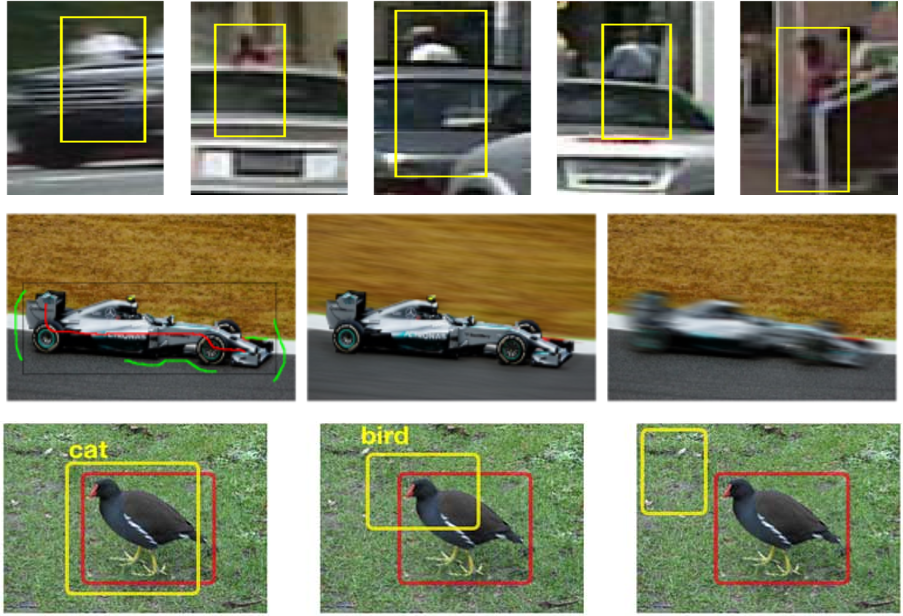}
	\caption{Several challenges exist for a robust tracking and detection system. (a) occlusions, (b) clutter detection, (c) motion blur. }
	\vspace{-0.3cm}
	\label{fig:det_chan}
\end{figure}
Secondly, CNN's shift covariance properties allow generating regional proposals from the area of maximum responses on object specific detection filters.

A common deep CNN-based video object tracking system consists of two parts of object detection and object displacement networks. R-CNN \cite{ren2015faster} network is the most commonly used backbone structures for object detection. Generally, a R-CNN network consists of two stages. In the first stage, a RPN network generates object likeness score and bounding box offset coordinates from a fixed number of predefined anchors on the output feature maps. In the second stage, plausible candidate regions of high object likeness score are pooled on another feature maps for refined coordinates offsets and object class score. For object displacement predictions, a correlation layer takes the features from Siamese network on reference frames and prediction frames for object displacement predictions. Several works \cite{kang2017t,han2016seq, zhu2017flow} refine the results by proposing multi-stages of regional proposal detection, feature propagation, objects tube linking and post-processing.  



The main challenge for robust tracking includes motion blur, partial occlusions and background clutters. This poses challenges for developing robust tracking algorithms by adding uncertainties for object state estimations. Here we show some typical challenges for robust detection and tracking in Fig. \ref{fig:det_chan}
In response to the increasing concerns about model robustness and generalizations without introducing extra cost, we see a resurgence of Bayesian models in recent years. Bayesian approaches use a probabilistic treatment of object appearances and states to deal with the uncertainties in prediction from models. However there are several challenges for taking existing works off-the-shelf for learning and making inference on network model outputs. Firstly, there are multiple objects appear and disappear in consecutive frames. Secondly, different objects linking possibilities exists for tracking objects across different frames. Finally, modeling the uncertainties of dynamic number of appeared object's states from a fixed R-CNN structure's outputs is still not well solved. In our work, we propose to address the first and the second problem by formulating a joint object dynamic over the distributions of a cascaded event of object appearing/disappearing, new object arriving and object associations. And we address the third problem by considering the R-CNN outputs as a clustered emission distributions from object's ground states over appearance scores, classification scores and location coordinates.

In this paper, we formulate the problem of multi-object detection and tracking in a fully probabilistic way. Our model is parameterized by tracking and detection neural networks. We take the original network structure from the original detect to track paper \cite{feichtenhofer2017detect} with minimum modification in notations and definitions.  Our formulation consists of a transition models for objects dynamics and emission model for object detection. We take neural network's outputs as our transition and emission distribution's parameters. Our probabilistic model is capable of handling the multi-object appearing and disappearing problem by incorporating an object appearance and association prior. In our model, we see tracking and detection as inferring objects states posteriors from network outputs. And the posterior inference is proposed by our particle filter based sampling algorithm. We use our particle filter algorithm to generate samples from an approximated family of distributions and later re-weigh and re-sample in accordance with our model formulations. Our sampled trajectories is capable of taking the relations cross all visible frames into account. Finally, we present a Variation Sequential Monte Carlo (VSMC) \cite{maddison2017filtering} method to learn our model from intermittent labeled frames. Our VSMC method trains on unlabeled consecutive frames by optimizing a tractable evidence lower bound (ELBO).  The ELBO is approximated from sampled labels generated by our particle filter algorithm. Our contribution in this paper are three folds.   
\begin{itemize} 
\item We give a deep neural network parameterized Bayesian formulation of a detection and tracking system.
\item We propose a particle filter sampling algorithm to make robust estimation on tracking object states from neural network outputs.
\item We present a weakly-supervised training algorithm for our model by VSMC methods. Our VSMC algorithm trains on both labeled frames and unlabeled consecutive frames with the pseudo-label generated by our sampling algorithm.  
\end{itemize}



\section{Related Work}
A comprehensive review of tracking and detection algorithm is beyond the scope of this paper. We review the works that are most related to our study.\\

\textbf{Object Detection and Tracking}
Currently, state-of-art object detection and tracking systems consist of multiple stages of regional proposal detection, feature propagation, object tube linking and post-processing. In the first stage is to extract regional proposal candidates. Representing works include 2D R-CNN network in frame-level box proposal \cite{han2016seq}, 3D R-CNN network in video-level tube proposal \cite{kang2017t} and feature propagation \cite{zhu2017flow}. In the second stage, detected objects are linked together to make a tracking prediction. Either direct detection box \cite{zhang2018integrated} or tracking displacement \cite{feichtenhofer2017detect} offset box are linked by bipartite graph algorithm \cite{frank2005kuhn} or Viterbi algorithm \cite{forney1973viterbi}. In the final stage, suppression methods are used for removing duplications and false positives. 
\vspace{-1.0em}\\

\textbf{Correlation Filter and Siamese Network} Correlation filters have recently been introduced into visual tracking and shown to achieve high speed as well as robust performance \cite{bolme2010visual,valmadre2017end}. Siamese network with triplet loss has also shown advantages in clustering similar object samples, which has been commonly used in representation learning\cite{ji2021improving}. Bo \cite{li2018high} proposes a Siamese-RPN network that are trained end-to-end to regress tracking template locations. Bo \cite{li2019siamrpn++} improves the performance by using a ResNet-50 backboned CNN and taking RPN predictions by aggregating multi-layer outputs. Zhipeng \cite{zhang2019deeper} proposes a deeper and wider Siamese network by adding multiple crops to remove padding effects. Instead of cropping the objects on template images, Feichtenhofer \cite{feichtenhofer2017detect} uses ROI poolings on correlation feature maps for object displacement prediction from template to target. 
\vspace{-1.0em}\\

\textbf{Bayesian Model for Tracking and Detection} Bayesian methods have been applied to object detection and tracking for its robust performance to deal with object state uncertainties. Tianzhu \cite{zhang2017multi} combines the power of particle filter and correlation filter for robust tracking objects on image and video frames.  They solve a correlation filter in dual space by accelerated proximal gradient method. And then they propose a particle filter tracker to generate particles using transition model, apply proposed correlation filter to shift it to a stable location and reweights the sample using the filtered responses. However their correlation filter is based on the shallow features in Fourier domains. Ali \cite{harakeh2019bayesod} combines the power of Bayesian inference and CNN for dealing with uncertainties of deep neural network's observations. They assume a Gaussian prior on object location and  Dirichlet prior on object categories. Xinshuo \cite{Weng2019_3dmot} proposes 3D object detection baseline. They use Gaussian distribution to formulate the object location uncertainties and use Kalman Filter to infer the trajectories of each single objects. \vspace{0.6em}\\

\section{A Bayesian Formulation of Object Detection and Tracking System}
In this section, we give a bayesian formulation of object detection and tracking, We consider the joint process of object detection and tracking as a HMM, where the distribution of object states cross frames are modeled as hidden states. We view tracking as the transition between neighboring hidden states and detection as emission from hidden states to a noisy visible states observed by R-FCN network outputs. We would show that in the case of supervised learning, the traditional tracking and detection loss takes a similar form of the maximum likelihood function in our formulated model. 
\subsection{Definition and Notations}
We denote the state's of a object indexed at $i$ at frame $t$ as $\mathcal{B}_{i,t}$. And $\mathcal{B}_{i,t}$ is defined as as a tuple of three variables.
\begin{equation}
    \mathcal{B}_{i,t}\triangleq  \{ L_{i,t}, \mathcal{C}_{i} \}
\end{equation}
where $L_{i,t}$ denotes the vector of bounding box's locations, $\mathcal{C}_{i}$ denotes the category of the object.

The category of the object $\mathcal{B}_{i,t}$ keeps unchanged cross all the frames $\mathcal{B}_{i,t}$ appears. And the $L_{i,t}$ location shifts between frames. We model the transitional distributions of object i between frame $t$ and $t+1$ as a neural network parameterized Gaussian distributions. 
\begin{equation}
\label{eq1}
p_g(L_{i,t+1}|L_{i,t})\sim \mathcal{N}(g^\phi_\mu(\mathcal{B}_{i,t}), g^\phi_\sigma(\mathcal{B}_{i,t}))
\end{equation}
where $g^\phi_\mu(\mathcal{B}_{i,t}), g^\phi_\sigma(\mathcal{B}_{i,t})$ is given by the RoI output's from the deep correlational kernel network in Feichtenhofer\cite{feichtenhofer2017detect}.

And at each frame $t$, the R-FCN gives M anchored observations as $[\hat{\mathcal{B}}_{1,t},...\hat{\mathcal{B}}_{M,t}]$. Each anchored observations is either anchored with a true object $\mathcal{B}_{i,t}$ with above definition or anchored with a clutter observations (false positive). We make a simplified assumption that each box output corresponds to a unique objects in the ground truth state. Actually, this is the most widely adopt assumptions in traditional R-FCN training and inference.
In our simplified assumptions, we introduce an anchor to objects variable $\mathbf{u}$ which is given by
\begin{equation}
        u_i=
     k \in \{1,2 ... K_t\} 
\end{equation}
where the R-FCN's anchored observation $\hat{\mathcal{B}}_{i,t}$ observation is clustered around objects $\mathcal{B}_{k,t}$. In our model, the emission distributions from clusters $\mathcal{B}_{k,t}$ to anchored observations also follows a network parameterized Gaussian distributions
\begin{equation}
\label{eq2}
    p_f(\hat{L}_{i,t}|L_{k,t}) \sim \mathcal{N}(L_{k,t}, f_\theta(\mathcal{B}_{k,t})) 
\end{equation}
To distinguish between ground truth observation and clutter observation, we use another binary variable $\mathcal{E}_{i,t}$ which is given by
\begin{equation}
    \mathcal{E}_{i,t} =
    \begin{cases}
    0     & \quad \mathcal{B}_{k,t} \text{is associated with one real object} \\
    1  & \quad \mathcal{B}_{k,t} \text{is a clutter observations}
    \end{cases}
\end{equation}

\subsection{Tracking as Object Dynamics and Association}
We assume that there are $K_t$ objects appears at frame $t$. At next frame, each object remain/disappears independently at death probability $\lambda_D$ with $\hat{K}_t$ remaining objects and $\Delta K_t$ new coming target arrive at the rate of $\lambda_{L}$ following Poisson distributions. The whole object appearance dynamics could be written as
\begin{equation}
p(\hat{K}_t, \Delta K_t)= {\lambda_D}^{ K_t - \hat{K}_t}{(1-\lambda_D)}^{\hat{K}_t} \frac{\lambda_{L}^{\Delta K_t}e^{-\lambda_{L}}}{\Delta K_t!}
\end{equation}
Considering all of the objects are observed in an unknown order, we formulate the distribution of a reordering of $\hat{K}_{t+1}+\Delta K_t$ objects by introducing a measurement to target association(M$\to$T) hypothesis as $\lambda =(\mathbf{r},\hat{K}_{t+1},\hat{K}_{t+1}+\Delta K_t)$. And the element of association vector is defined as $\mathbf{r}=(r_1,r_2,r_3,...r_{\hat{K}_{t+1}})$ which is given by
\begin{equation}
    r_j = k \in \{1,2,3...\hat{K}_{t+1}+\Delta K_t\}
\end{equation}
where the j's object at the frame $t$ is associated with k's object at frame $t+1$.
And we also assign a uniform prior for the association vector $\mathbf{r}$
\begin{equation}
    p(\mathbf{r})= 
    \binom {\hat{K}_{t}+\Delta K_t}{\hat{K}_{t}}^{-1}
\end{equation}
The newly appeared object $\mathcal{B}_{j,t+1}$ follows a Gaussian prior on object locations $L_{j,t+1}$ and a uniform prior on object categories $\mathcal{C}_{j}$
\begin{align}
    \label{eqn0}
    p_0(L_{j,t+1}) & \sim \mathcal{N}(\mathbf{\mu}_0, \Sigma_0) \nonumber \\
    p_0(\mathcal{C}_{j}) & \sim F(k, 0, K-1)
\end{align}
For existing object $\mathcal{B}_{j,t+1}$, its location $L_{j,t+1}$ is updated by following transitional distribution defined in Equation.\ref{eq1} which is parameterized by tracking network. And the categories $\mathcal{C}_{j}$ keeps the same with its previous state. 

\begin{figure}[htb]
	\centering

	\includegraphics[width=\linewidth]{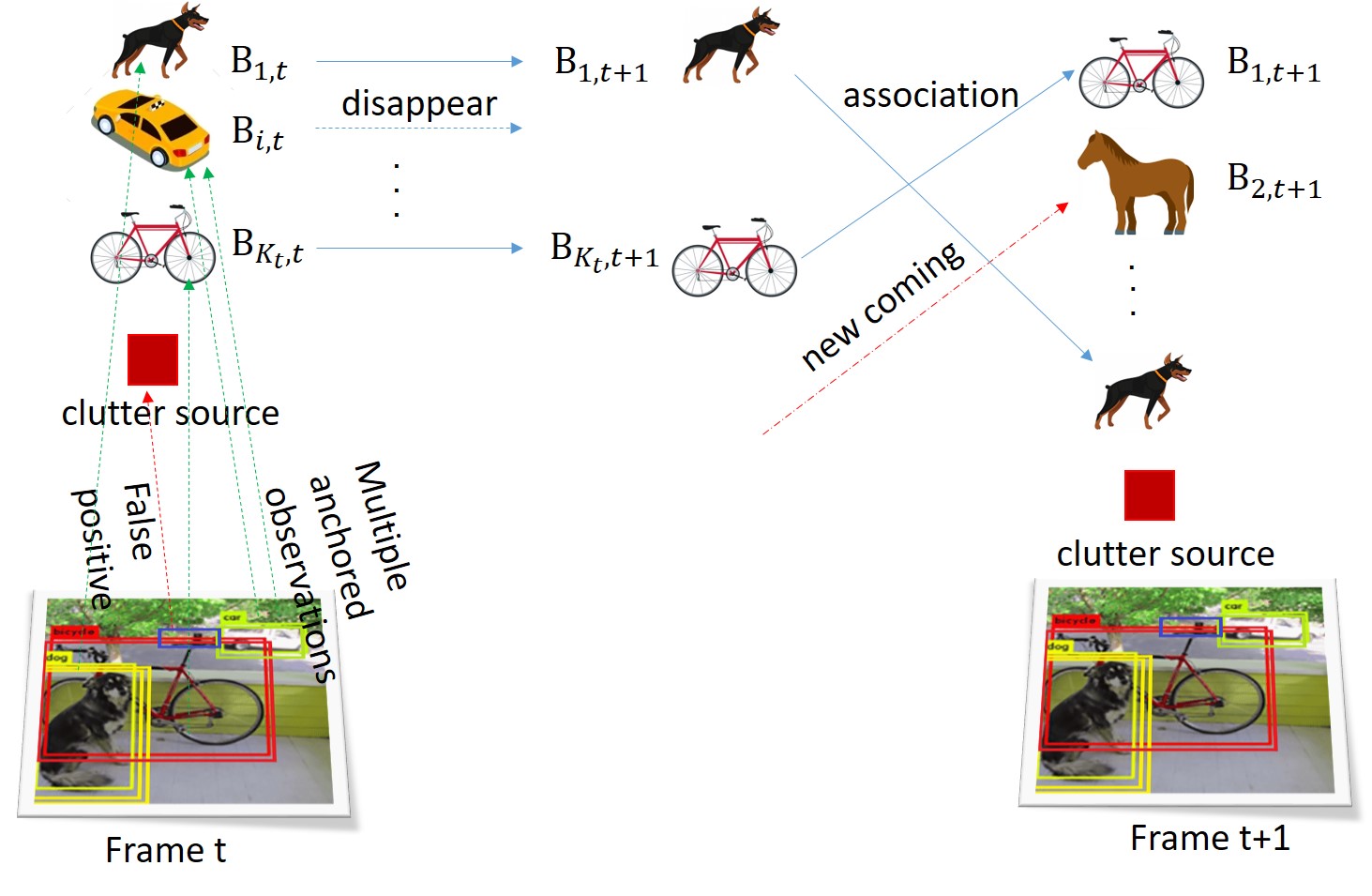}
	\caption{The whole object dynamics in our model. Multiple objects appear/disappear and observed in unknown sequence.} \label{fig:formulation1}
\end{figure}

With the above definition, the tracking process is modeled as the joint transition probability of object dynamics and associations  
\begin{align*}
&p(\{ \mathcal{B}_{1,t+1}...\mathcal{B}_{\hat{K}_{t+1}+\Delta K_t, t+1}\}, \mathbf{r}|\{\mathcal{B}_{1,t}...\mathcal{B}_{ K_t,t} \}) \\
=& p(\mathbf{r})p(\hat{K}_t, \Delta K_t)\\
&\prod_{i=1}^{\hat{K}_t}p_g(L_{r_i,t+1}|L_{i,t}))I(\mathcal{C}_{r_i,t+1}|\mathcal{C}_{i,t})\prod_{i\notin \mathbf{r}}p_0(\mathcal{B}_{i})p_0(\mathcal{C}_{i})
\end{align*}
where $I(\cdot)$ is an indicator function where $I(\mathcal{C}_{r_i,t+1}|\mathcal{C}_{i,t})=1$ for $\mathcal{C}_{r_i,t+1}=\mathcal{C}_{i,t}$ and $I(\mathcal{C}_{r_i,t+1}|\mathcal{C}_{i,t})=0$ for $\mathcal{C}_{r_i,t+1} \neq \mathcal{C}_{i,t}$ 
The whole object dynamics is shown in Fig. \ref{fig:formulation1}

\subsection{Detection as Object Emission} 
We consider the emission distribution from a real object $\mathcal{B}_{i,t}$ to its anchored observations $\hat{\mathcal{B}}_{k,t}$ from R-FCN outputs. Recall the properties of a R-FCN, its emissions from the final layer of feature maps consists of three parts as object appearance score $e_{k,t}$, object classification score $\mathcal{K}_{k_t}$ and object location coordinates $L_{k,t}$. \\
Instead of viewing the probability output as a distribution over object's existence and categories directly, we see the R-FCN's's a categorial digits outputs as a categorical distribution conditioned on the real category $\mathcal{C}_{i}$ of its belonging object. And in our formulation $\mathcal{K}_{k_t}$ follows a Dirichlet distribution with $\alpha$ conditioned on $\mathcal{C}_{i}$
\begin{equation}
p(\mathcal{\hat{K}}_{i,t}|\mathcal{C}_{i}) \sim  \textit {Dir}(\alpha_1,......\alpha_K)
\end{equation}
in our formulation $\alpha_1,......\alpha_K$ is set as
\begin{equation}
    \alpha_k =
    \begin{cases}
    \alpha +1      & \quad \text{if } k = \mathcal{C}_{i} \\
    1  & \quad \text{if } k \neq \mathcal{C}_{i}
    \end{cases}
\end{equation}
Similarly, we treat the object appearance score $e_{k,t}$ as a categorical distribution in a beta distributions conditioned on $\mathcal{E}_{i,j}$ indicating whether the $\hat{\mathcal{B}}_{k,t}$ is anchored around a real object or a clutter observations.  
\begin{equation}
p(\hat{e}^M_{j,t}|\mathcal{E}_{i,t}) \sim  \textit {Beta}(\alpha_0,\alpha_1)
\end{equation}
 $\alpha_0,......\alpha_1$ is set as
 \begin{equation}
    \alpha_0,  \alpha_1=
    \begin{cases}
    \alpha+1, 1      & \quad \text{if }  \mathcal{E}_{i,t} = 0 \\
     1 , \alpha+1   & \quad \text{if }  \mathcal{E}_{i,t} = 1
    \end{cases}
\end{equation} 
As for $\hat{\mathcal{B}}_{k,t}$ is anchored around real object $\mathcal{B}_{i,t}$,
\begin{equation}
    p(\hat{e}_{j,t}|\mathcal{E}_{i,t}) \sim  \textit {Beta}(1 , \alpha+1)
\end{equation}
For object location coordinates outputs $\hat{L}_{k,t}$, we consider $\hat{L}_{k,t}$ as a noisy observation of object's real locations $L_{i,t}$. $\hat{L}_{k,t}$ follows a Gaussian distribution given by Eq. (\ref{eq2}) with its mean as $L_{i,t}$ and variance as $f_\theta(\mathcal{B}_{k,t}$. In our work, we do not treat $f_\theta(\mathcal{B}_{k,t})$ as a direct output by R-FCN network. Instead, We follow the line with \cite{harakeh2019bayesod} by viewing covariance as combined contributions of model uncertainty and prediction uncertainty. Similar with their work, we omit the model uncertainty and use prediction uncertainty $\Sigma(\mathcal{B}_{i,t})$ as an approximation for $f_\theta(\mathcal{B}_{k,t})$. The prediction uncertainty $\Sigma(\mathcal{B}_{i,t})$ is taking as the covariance of all output coordinate predictions $\hat{L}_{k,t}$ for all anchors around $\mathcal{B}_{k,t}$.
\begin{align}
\label{meanbox}
    \mu(\mathcal{B}_{i,t}) &= \frac{1}{M_i}\sum_{u_j=i}f^\theta(\mathcal{\hat{B}}_{j,t})\\
    \Sigma(\mathcal{B}_{i,t})& \approx \frac{\alpha}{M_i}(\sum_{u_j=i}f^\theta(\mathcal{\hat{B}}_{j,t})f^\theta(\mathcal{\hat{B}}_{j,t})^T)-\alpha \mu(\mathcal{B}_{i,t})\mu(\mathcal{B}_{i,t})^T
\end{align}
For anchored observation $\hat{\mathcal{B}}_{k,t}$ that associates with a clutter $\hat{\mathcal{B}}^0_{i,t}$, we assume a Gaussian prior for clutter location and uniform prior on clutter categories in the same form as Eq. (\ref{eqn0}).

\begin{figure}[htb]
	\centering

	\includegraphics[width=\linewidth]{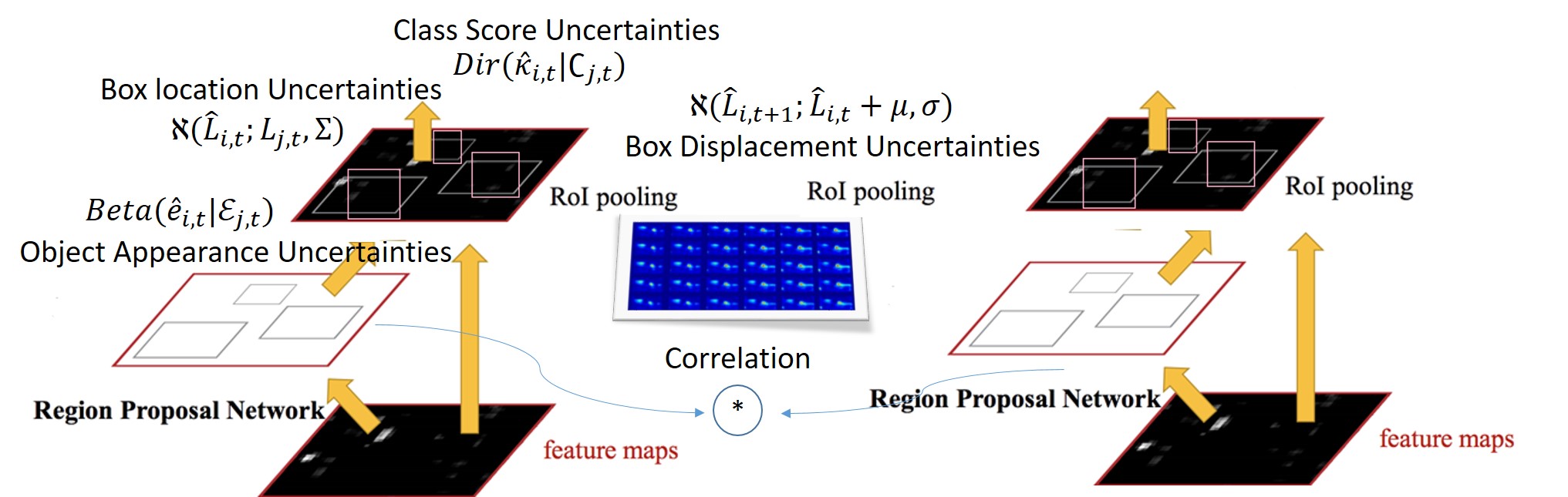}
	\caption{Out Bayesian view of traditional tracking and detection parameterized Output} \label{fig:bayesiannet}
\end{figure}

With the above definition, the joint emission probability is given by
\begin{align*}
    &p(\{\mathcal{\hat{B}}_{1,t}...\mathcal{\hat{B}}_{ M,t}\}|\{\mathcal{B}_{1,t}...\mathcal{B}_{ K_t,t}\})\\
    &=\Gamma\prod_{i\in (\hat{E}_{i,t}=1)}\hspace{-0.8em}\hat{e}_{j,t}^\alpha\mathcal{N}(f^\theta(\hat{L}_{i,t});L_{u_i,t},f_\theta(\mathcal{B}_{k,t}))c_{i}^\alpha \\ 
    &\sum_{L_{u_i,t}}\sum_{C_{u_i}}\prod_{i\in (\hat{E}_{i,t}=0)}\hspace{-0.8em}(1-\hat{e}_{j,t})^\alpha \mathcal{N}(f^\theta(\hat{L}_{i,t});L_{u_i,t},f_\theta(\mathcal{B}_{k,t}))c_{i}^\alpha\\ &    \hspace{0.6em}p_0(L_{j,t})p_0(C_{u_i})  
\end{align*}
where $c_{i}=\hat{K}_{i,t}(\mathcal{C}_{u_i})$ is the categorical score for its associated object's class.
Because we do not infer on its location and categories for negative anchors, we take the marginalized distribution for categorical and location output score for negative anchors. The whole R-FCN detection process is shown in Fig. \ref{fig:bayesiannet}

\subsection{Traditional D\&T Loss as Model Likelihood}
\label{supervise}
In the case of supervised learning, we have annotations of all hidden states. And all the visual states and distribution are parameterized by the network outputs. All the association variable is also annotated for tracking. We train the network by taking the log likelihood of joint probability as
\begin{align}
    \mathcal{L}& = \sum_{t=1}^{T}\log p(\{\mathcal{\hat{B}}_{1,t}...\mathcal{\hat{B}}_{ M,t}\}|\{\mathcal{B}_{1,t}...\mathcal{B}_{ K_t,t}\})\nonumber\\
    & + \sum_{t=1}^{T-1}\log p(\{ \mathcal{B}_{1,t+1}...\mathcal{B}_{\hat{K}_{t+1}+\Delta K_t, t+1}\}, \mathbf{r}_t|\{\mathcal{B}_{1,t}...\mathcal{B}_{ K_t,t} \})\nonumber\\
    &=\hspace{-2.5em} \sum_{i\in\{r_{i}\in\{1:K_{t+1}\}\},t}\hspace{-2.5em}\frac{1}{2}||L_{r_i,t+1}-g^\phi_\mu(\mathcal{B}_{i,t})||_{g^\phi_\sigma(\mathcal{B}_{i,t})^{-1}} + \frac{1}{2}\log(g^\phi_\sigma(\mathcal{B}_{i,t}))\nonumber\\
    &+ \sum_{i=1,t}^{i\leq M}\frac{1}{2}|||\hat{L}_{i,t}-L_{u_i,t}||_{\Sigma^{-1}} + \frac{1}{2}\log \det({\Sigma^{-1}(\mathcal{B}_{i,t})})\nonumber\\
    &+ H(e_{u_i}, \mathcal{E}_{u_i}) 
    + H(\mathcal{K}_{i,t},\mathcal{C}_{u_i})
    + const \nonumber\\
    & = \mathcal{L}_{track} + \mathcal{L}_{detect} +const
    \label{supervise_loss}
\end{align}
where $\mathcal{L}_{track}$ and $\mathcal{L}_{detect}$ is our commonly used training loss for deep correlational kernel tracking network(or Siamese Network) and R-FCN detection network. The only difference is that we introduce a $g^\phi_\sigma(\mathcal{B}_{i,t})$ as a parameterized network outputs indicating the independent variance for tracking coordinate prediction. And we put the term $\frac{1}{2}||L_{r_i,t+1}-g^\phi_\mu(\mathcal{B}_{i,t})||_{g^\phi_\sigma(\mathcal{B}_{i,t})^{-1}} + \frac{1}{2}\log(g^\phi_\sigma(\mathcal{B}_{i,t}))$ for joint training of tracking mean and variance. We use traditional detection loss for training detection network and omits the term $\frac{1}{2}\log \det({\Sigma^{-1}(\mathcal{B}_{i,t})})$ in our training. Here we also omit the term of marginal distribution of location and categories for clutter anchors.

\section{A Particle Filter Object State Estimation Algorithm For Robust Tracking}
\label{pfinfer}
With our formulation in previous section, our robust tracking is by inferring the posterior of all objects states and their associations cross frames from the observations on R-FCN and tracking dynamics on tracking network. As our probabilistic formulation is straightforward for joint probability verification but less straightforward for sampling posterior, directly inferring the joint states of objects and their associations would be intractable. An analytical solution would require traversing all the possible object appearing and association states which is impractical for implementation. Here we give a particle filter based sampling solution by sampling from an approximated family of straightforward sampling distributions. 

\textbf{Initial Sample on R-FCN Detectors}\hspace{1em} 
Assume that R-FCN outputs M anchored prediction of bounding box state
$\{\mathcal{\hat{B}}_{1,t}...\mathcal{\hat{B}}_{ M,t}\}$ at frame $t$. And we sample $\{\mathcal{B}_{1,t}...\mathcal{B}_{ K_{t},t}\}$ according the the posterior
\begin{equation*}
r_1(\{\mathcal{B}_{1,t}...\mathcal{B}_{K_{t},t}\}) \propto  p(\{\mathcal{\hat{B}}_{1,t}...\mathcal{\hat{B}}_{ M,t}\}|\{\mathcal{B}_{1,t}...\mathcal{B}_{K_{t},t}\})
\end{equation*}
And we sample $r_1(\{\mathcal{B}_{1,t}...\mathcal{B}_{K_{t},t}\})$in the following way(the proof would be given in the later part of this section). First, per-anchor outputs from the neural network are clustered in a similar way with NMS. Greedy clustering is performed using the output category score by choosing the anchor with the highest non-background score as the cluster center, adding any anchor with an intersection over union(IOU) greater than 0.5 to the cluster, and eliminating all members in the cluster from the original updated anchor set. This process is repeated until all the anchors are assigned to clusters. By seeing each cluster as a object candidate, we have candidates objects of $\{\tilde{\mathcal{B}}_{1,t}...\tilde{\mathcal{B}}_{K_{t},t}\}$. Then we sample the objects are their states in the following way.
\begin{enumerate}[Step 1:]
\item Include object $\tilde{\mathcal{B}}_{i,t}$ with probability $p_i$
\begin{equation*}
p_i =\prod_{j\in (u_j=i)}(\hat{e}_j)^\alpha/(\hspace{-0.8em}\prod_{j\in (u_j=i)}(\hat{e}_j)^\alpha+\hspace{-0.8em}\prod_{j\in u_j=i}(1-\hat{e}_j)^\alpha)
\end{equation*}
\item Assign initial box state $\mu(\mathcal{B}_{i,t})$ by Eq. (\ref{meanbox})
\item Assign a new tracking id for each  $\mathcal{B}_{i,t}$. 
\item Sample object class $\mathcal{C}_i \sim \text{Categorical}(c_i)$
\begin{equation*}
c_i=\prod_{j\in(u_j=i)}\mathcal{K}_{j,k}^\alpha/\sum_{k'} \prod_{j\in(u_j=i)}\mathcal{K}_{j,k'}^\alpha
\end{equation*}
\end{enumerate}
\vspace{-0.5em}

\textbf{Data Association}\hspace{1em} After we include our sampled objects and their states, we associate our generated initial objects $\{\mathcal{B}_{1,t}...\mathcal{B}_{\tilde{K}_{t},t}\}^l $ to the objects in their ancestor particles $\{\mathcal{B}_{1,t-1}...\mathcal{B}_{ K_{t-1},t-1}\}^{a_{t-1}^{l}}$. To match these two sets of objects, we first construct a affinity matrix ${J}$ with dimension of $\tilde{K}_{t}\times K_{t-1}$. We assign a score to the elements of $J$ using the summation of IoU between objects and their class scores
\begin{equation}
    J_{i,j} = \log IoU(g_{\phi}^{\mu}(\mathcal{B}_{i,t-1}^{a_{t-1}^{l}}), \mu(\mathcal{B}_{j,t})) + \log \mathcal{K}_{j,\mathcal{C}_i}
\end{equation}
Then we apply Hungarian algorithm to solve the bipartite graph matching problem in polynomial time. In addition, we reject matching objects $\mathcal{B}_{i,t-1}$ with $\mathcal{B}_{j,t-1}$ when $\mathcal{B}_{j,t}$ when their IoU is below then a given threshold of $IoU_{min}$. Without loss of generality, we assume to output a set of matched objects $\{\hat{\mathcal{B}}_{1,t}...\mathcal{B}_{\hat{K}_{t},t}\}^l$ in frame $t$ with their ancestor $\{\hat{\mathcal{B}}_{r_1,t-1}...\mathcal{B}_{r_{\hat{K}_{t}},t-1}\}^{a_{t-1}^{l}}$ in frame $t-1$, along with the unmatched objects $\{\mathcal{B}_{\hat{K}_{t}+1,t}...\mathcal{B}_{\tilde{K}_{t},t}\}^l$. \vspace{-0.5em}\\ 

\textbf{Independent Proposal Update for Tracking Object}
After we associate our sampled objects with their ancestors, we update the state of each sampled objects independently by sampling their posterior conditioned on their ancestors. Prior on the transition probability of matched objects, we sample each object's bounding box state from their corresponding posterior. We use this proposal distribution to minimise the variance of the importance weights
\begin{align}
    r_2(\mathcal{B}_{i,t}) & \propto p(\mathcal{B}_{i,t}|\mathcal{B}_{r_i^{-1},t-1}^{a_{t-1}^{l}})\prod_{j\in u_j=i}p(\hat{\mathcal{B}}_{j,t}|\mathcal{B}_{i,t}) \nonumber \\
    \label{eqr2}
    & = \mathcal{N}(\mu'(\mathcal{B}_{i,t}),\Sigma'(\mathcal{B}_{i,t}))
\end{align} 
where the sufficient statistics of proposal in Eq. (\ref{eqr2}) can be estimated in closed form as:
\begin{align}
    \Sigma'(\mathcal{B}_{i,t}) &= (\mathrm{diag}^{-1}(g_{\phi}^{\sigma}(\mathcal{B}^{a_{t-1}^{l}}_{r_i^{-1},t-1}))+M_i\Sigma^{-1}(\mathcal{B}_{i,t}))^{-1}\\
    \mu'(\mathcal{B}_{i,t}) &= \Sigma'((g_{\phi}^{\sigma}(\mathcal{B}^{a_{t-1}^{l}}_{r_i^{-1},t-1}))^{-1}\otimes g_{\phi}^{\mu}(\mathcal{B}^{a_{t-1}^{l}}_{r_i^{-1},t-1}) \nonumber\\
    &+M_i\Sigma^{-1}(\mathcal{B}_{i,t})\mu(\mathcal{B}_{i,t}))
\end{align}
where $M_i$ is the number of anchored observations for object $\mathcal{B}_{i,t}$ in R-FCN. 

Similarly, for unmatched objects $i'$, 
 $\mu'_0(\mathcal{B}_{i',t}),\Sigma'_0(\mathcal{B}_{i',t})$ could be given by
\begin{align}
    \Sigma_0'(\mathcal{B}_{i',t}) &= (\Sigma^{-1}_0+M_{i'}\Sigma^{-1}(\mathcal{B}_{i',t}))^{-1}\\
    \mu'_0(\mathcal{B}_{i',t}) &= \Sigma'(\Sigma^{-1}_0\mu_0 +M_{i'}\Sigma^{-1}(\mathcal{B}_{i',t})\mu(\mathcal{B}_{i',t}))
\end{align}
Finally, we keep the sampled class $\mathcal{C}_{i'}$ for unmatched objects ${i'}$. And we discard the sampled class for matched object $i$, and assign $\mathcal{C}_{i}$ with the class of its matched ancestor objects $\mathcal{C}_{r^{-1}_i}^{a_{t-1}^l}$. Similarly, we keep the tracking id for unmatched objects and assign matched objects $\mathcal{B}_{i,t}$ with an id of $id(\mathcal{B}_{r_i^{-1},t-1}^{a_{t-1}^l})$\vspace{-0.5em}\\ 

\begin{theorem}
\label{theorem1}
The importance weight $w_t^l$ of our sampled objects state $\{\mathcal{B}_{1,t}...\mathcal{B}_{\tilde{K}_{t},t}\}^l$ from its ancestor state $\{\mathcal{B}_{1,t-1}...\mathcal{B}_{ K_{t-1},t-1}\}^{a_{t-1}^{l}}$ is given by
\vspace{-1.5em}
\begin{align}
    w_t^l \propto& \nonumber \underbrace{(1-\lambda_D)^{K_{t-1}-\hat{K}_t}\lambda_D^{\hat{K}_t}}_{\text{Object Appearance}} \underbrace{p((\tilde{K}_t-\hat{K}_t);\lambda_{L})}_{\text{New Object Arriving}}\underbrace{\binom {\tilde{K}_t}{\hat{K}_t}^{-1}}_{\text{Association Prior}}\\
    &e^{-\frac{1}{2}R_D}\tau_D\hspace{-1em}\underbrace{\prod_{i\in \mathbf{r}}\mathcal{K}_{i,\mathcal{C}_{  r^{-1}_i}^{a_{t-1}^l}}}_{\text{Tracking Class Probability}}
\end{align}
\vspace{-0.4em}
where is $R_D$ is given by
\vspace{-0.4em}
\begin{align}
R_D & = \sum_{i\in \mathbf{r}}\underbrace{g_{\phi}^{\mu}(\mathcal{B}^{a_{t-1}^{l}}_{r_i^{-1},t-1})^{\circ2}\odot((g_{\phi}^{\sigma}(\mathcal{B}^{a_{t-1}^{l}}_{r_i^{-1},t-1})^{-1})}_{\text{Tracking Transition Prior}} \nonumber\\
    & 
    - \underbrace{\mu_0^T\Sigma^{-1}_0\mu_0}_{\text{New Object Prior}} - \underbrace{\mu'(\mathcal{B}_{i,t})^T\Sigma'(\mathcal{B}_{i,t})^{-1}\mu'(\mathcal{B}_{i,t})}_{\text{Matched Location Posterior}} \\
    & + \underbrace{\mu'_{0}(\mathcal{B}_{i,t})^T\Sigma'_{0}(\mathcal{B}_{i,t})^{-1}\mu'_{0}(\mathcal{B}_{i,t})}_{\text{Unmatched Location Posterior}}
\end{align}
$\tau_D$ is given by
\begin{align}
 \tau_D & = \prod_{i\in \mathbf{r}}|g_{\phi}^{\sigma}(\mathcal{B}^{a_{t-1}^{l}}_{r_i^{-1},t-1})|^{-1/2}_{\otimes} \det(\Sigma_0)^{1/2}\nonumber\\
  &\det(\Sigma'(\mathcal{B}_{i,t})^{-1})^{1/2}\det(\Sigma'_{0}(\mathcal{B}_{i,t})^{-1})^{-1/2}  
\end{align}
\end{theorem}
\vspace{-1em}
\section{VSMC Algorithm For Semi-Supervised Model Learning}
In this section, we describe our model training algorithm in the case of semi-supervised settings. In such settings, we only have the labeled bounding box on frame $t$. In the neighboring frames $\{t-T:t-1\}$ and $\{t+1:t+T\}$, only visual input is given without labeled bounding box. Different from our supervised learning algorithm in section\ref{supervise}, training on the joint distribution for the unlabled frames is not feasible. Instead, our objective function is set as log likelihood of the marginal distributions where the unseen annotations are marginalized.
\begin{align}
\mathcal{L}& = \log p(\{\underbrace{\mathcal{\hat{B}}_{1:M,t-T}\},..\{\mathcal{\hat{B}}_{1:M,t}\},\{\mathcal{\hat{B}}_{1:M,t+T}\}}_{\text{Network Predictions from Data}}, \hspace{-0.6em}\underbrace{\{\mathcal{B}_{1:K_t,t}\}}_{\text{Labels for frame} t}\hspace{-0.6em})\nonumber\\
&= \underbrace{\log p(\{\mathcal{\hat{B}}_{1:M,t}\} , \{\mathcal{B}_{1:K_t,t}\})}_{\mathcal{L}_{detect}   \text{at frame} t}\nonumber \\
&+ \underbrace{\log p(\{\mathcal{\hat{B}}_{1:M,t+1}\}, ...\{\mathcal{\hat{B}}_{1:M,t+T}\}|\{\mathcal{B}_{1:K_t,t}\})}_{\text{Forward Marginal Likelihood}}\nonumber \\
\label{semiloss}
&+ \underbrace{\log p(\{\mathcal{\hat{B}}_{1:M,t-T}\}, ..\{\mathcal{\hat{B}}_{1:M,t-1}\}|\{\mathcal{B}_{1:K_t,t}\})}_{\text{Backward Marginal Likelihood}} \nonumber \\
= &\sum_{t'=-T,t'\neq 0}^{T}\log p(\{\mathcal{\hat{B}}_{1:M,t+t'}\}|\{\mathcal{B}_{1:K_{t+t'},t+t'}\})\nonumber\\ 
    & + \hspace{1em}\log p(\{\mathcal{B}_{1:K_{t+t'+1},t+t'}\}|\{\mathcal{B}_{1:K_{t+t'},t+t'-1}\}) \nonumber \\
& + \log p(\{\mathcal{\hat{B}}_{1:M,t}\} , \{\mathcal{B}_{1:K_t,t}\})    
\end{align}
Eq. (\ref{semiloss}) is derived from the property of HMM. The total loss objective could be decomposed into three terms. The first term $\log p(\{\mathcal{\hat{B}}_{1:M,t}\} , \{\mathcal{B}_{1:K_t,t}\})$ is the supervised detection loss on the same form as the traditional R-FCN detection loss. And the second term is the marginal likelihood of forward predictions on frame $\{t+1:t+T\}$. The third term is the marginal likelihood of backward prediction on frame $\{t-T:t-1\}$. Both marginal likelihood for forward and backward predictions are conditioned on frame $t$'s annotation $\{\mathcal{B}_{1:K_t,t}\}$ as a weakly supervised training signal for the unlabeled neighbor frames. As the backward likelihood could be factorized in the same structure of forward likelihood. We only derive the variational bound for forward likelihood in later part of the section. For the semi-supervised term, training directly is intractable, we use the following surrogate ELBO as a substitution
\vspace{-1em}
\begin{equation}
\vspace{-1em}
    \tilde{\mathcal{L}} = \sum_{t'=1}^{T} \mathbb{E}_{r_(\mathcal{B}_{t-1:t-t'}^{1:N},a_{1:t'-1}^{1:N};\lambda)}[\log (\frac{1}{N}\sum_{l=1}^{N}w_{t+t'}^l)]
\end{equation}

\begin{figure}[htb]
	\centering

	\includegraphics[width=\linewidth]{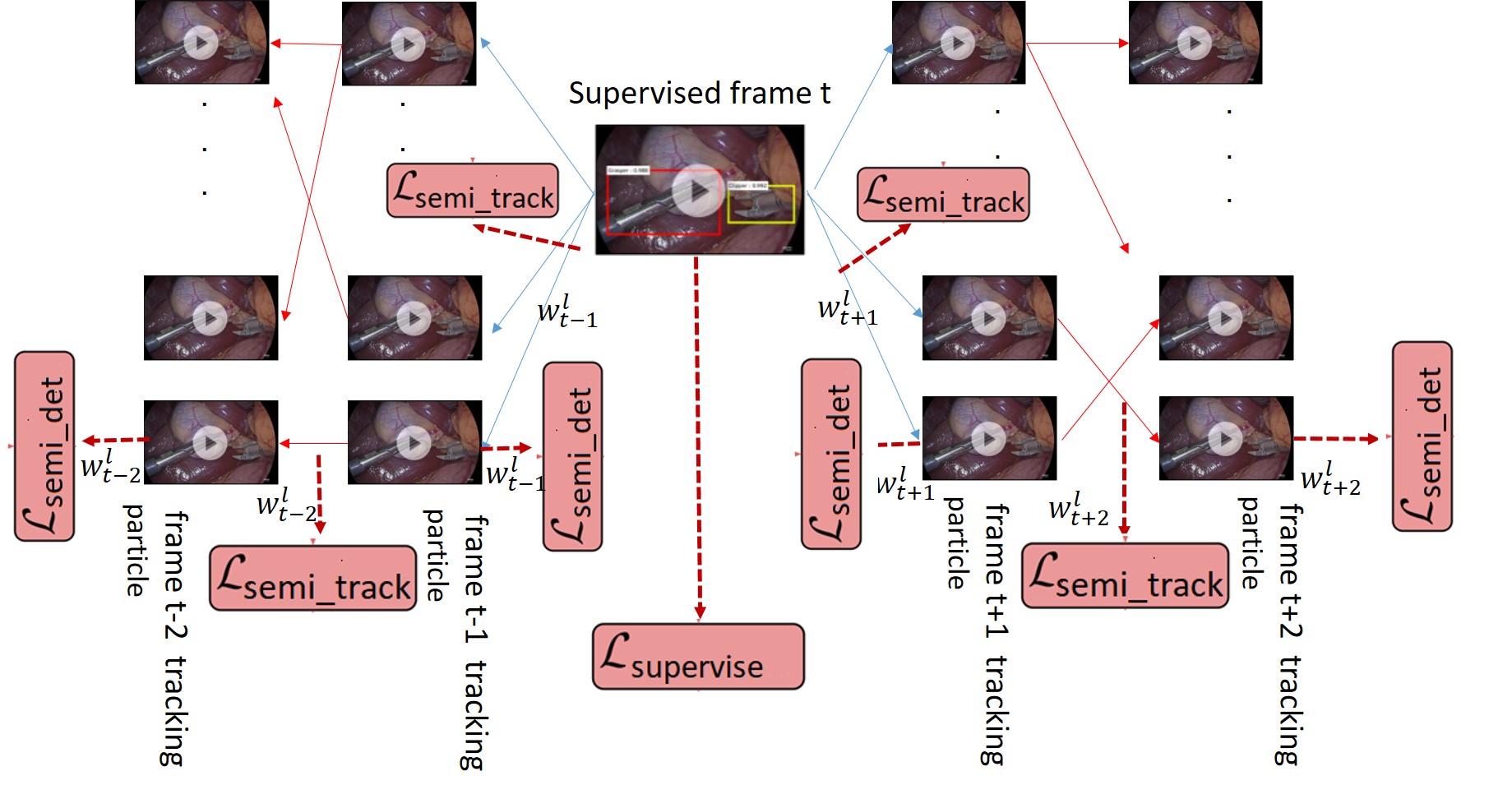}
	\caption{Our semi-supervised learning algorithm takes the sum of supervised loss on labeled frame and semi-supervised loss on consecutive frames. Our loss takes the weighted sum cross all sampled particles on their weights. The loss term for each sampled particle consists of detection loss on its own frames and tracking loss from the previous frame on its sampled trajectory.} \label{fig:semi}
\end{figure}

For convenience of understanding the surrogate ELBO loss for semi-supervised learning of our tracking and detection model. We link the surrogate ELBO loss to our widely adopted tracking and detection loss. By taking the gradient, the surrogate loss could be rewritten as
\begin{align}
     \nabla\tilde{\mathcal{L}} &= \sum_{t'=1}^{T} \mathbb{E}_{r_(\mathcal{B}_{t-1:t-t'}^{1:N},a_{1:t'-1}^{1:N};\lambda)}\nabla\log (\frac{1}{N}\sum_{l=1}^{N}w_{t+t'}^l)\nonumber\\
     & =\sum_{t'=1}^{T} \mathbb{E}_{r_(\mathcal{B}_{t-1:t-t'}^{1:N},a_{1:t'-1}^{l:N};\lambda)} \sum_{i=1}^{N}\hat{w}_{t+t'}^l\nabla\log (w_{t+t'}^l)\nonumber \\
     & \approx\hspace{-0.3em}\sum_{t'=1}^{T}\sum_{i=1}^{N}\hat{w}_{t+t'}^l\nabla\log (w_{t+t'}^l)\nonumber\\
     & \approx\hspace{-0.7em}\sum_{t'=t}^{t+T-1}\hspace{-0.3em}\sum_{l=1}^{N}\hat{w}_{t'+1}^l\nabla\log p(\{\mathcal{B}^l_{1:K_{t'+1},t'+1}\}|\{\mathcal{B}^l_{1:K_{t'},t'}\}^{a_{t'}^l}) \nonumber\\ 
     &+ \hat{w}_{t'+1}^l \nabla\log p(\mathcal{\hat{B}}^l_{1:M,t'+1}\}|\mathcal{B}^l_{1:K_{t'+1},t'+1}\})\nonumber\\
     &- \hat{w}_{t'+1}^l \nabla\log r(\mathcal{B}^l_{1:K_{t'+1},t'+1}; \lambda)\nonumber \\
     &\approx \hspace{-0.7em}\sum_{t'=t}^{t+T-1}\hspace{-0.3em}\sum_{l=1}^{N}\hat{w}_{t'+1}^l\nabla\mathcal{L}_{track}^{t'+1}(a_{t'}^l\to l) + \hat{w}_{t'+1}^i\nabla\mathcal{L}_{detect}^{t'+1}(l) \nonumber\\
     & - \hat{w}_{t'+1}^l \nabla\log r(\mathcal{B}^l_{1:K_{t'+1},t'+1}; \lambda)
\end{align}
where $\hat{w}_{t+t'}^l=w_{t+t'}^l/\sum_{l} w_{t+t'}^l$. 
The surrogate loss could be nicely decomposed into the weighted sum of tracking and detection loss over frame $\{t+1:t+T\}$ minus the log density of sampling distributions on sampled particle's importance weight. Built on the our particle filter sampling algorithm, our semi-supervised learning algorithm could be easily implemented on traditional tracking and detection loss. The only modification is that we include a set of sampled particles as a proxy for object annotation and take the weighted sum of their supervised loss by their importance weight. And this training step is iterated in the whole training process.To avoid introducing additional variations, we omit gradients of $\log r(\mathcal{B}_{1:K_{t'+1},t'+1}; \lambda)$. The detailed training loss is shown in Fig. \ref{fig:semi}.

\section{Experiments}
To show the effectiveness of incorporating uncertainty treatment in  tracking and detection comparing with traditional methods, we compare our performance with non-Bayesian baselines. Our evaluation is on two commonly used datasets. 

\textbf{ImageNet Video Object Detection Dataset (ILSVRC)} \cite{ILSVRC15} contains 30 classes in 3862 training and 555 validation videos. The objects have ground truth annotations of their bounding boxes and track IDs.

\textbf{M2Cai16-Tool-Locations Dataset} \cite{jin2018tool} extends the
M2Cai16-tool dataset. It contains 15 videos record at 25 fps of cholecystectomy procedures at the University Hospital of Strasbourg in France. Among those videos, 2532 frames are labeled under the supervision and spot-checking from a surgeon with medical devices including Grasper, Bipolar, Hook, Scissors, Clipper, Irrigator and Specimen Bag. 


To have a fair comparison of the Bayesian approach with the baseline, all the methods use the same structure of training and inference network and sharing training configurations in a much similar way. We only introduce an additional loss term for object transition's covariance matrix (only diagonal elements) in Eq. (\ref{supervise_loss}).
\subsection{Evaluation Metrics:}
Two evaluation metrics are used for the predication uncertainties quantification. We follow the commonly used benchmark for mAP thresholds at IoU value of 0.5. We also use the a probabilistic measurement \textbf{Probability Based Detection Quality (PDQ)} \cite{hall2018probability} to jointly quantify the bounding box location and categorical uncertainties for detection estimations. The PDQ score increases as the estimated distribution overlaps with both label's maximum likelihood and uncertainties. 

\subsection{Ablation Studies}
Our proposed method is compared against five different baselines to study the effectiveness of incorporating Bayesian formulation in each individual part of tracking and detection systems. In each baseline, none or partial Bayesian formulations are considered. We refer the five separate baseline methods as: Single R-FCN frame detector (Single R-FCN), Greedy R-FCN box linking (Greedy R-FCN), Greedy tracking offset R-FCN box linking (Greedy D$\&$T) and frame-wise Bayesian inference (Frame Bayesian) and Kalman filter for single object trajectory linking (Kalman-Link).

Single R-FCN takes Greedy Non-maximum Suppression (NMS) outputs directly from R-FCN, while Frame Bayesian infers object states from all cluster boxes (including suppressed and non-suppressed) only with frame-level priors. Greedy R-FCN links Single R-FCN makes predictions frame-by-frame with bipartite matching on IoU score. Greedy D$\&$T adds a tracking estimation part from Single R-FCN and links Single R-FCN predictions on tracking offset IoU scores. Kalman-Link links object by bipartite matching and updates box locations by Kalman filter with the trajectories of matched objects. 

Table \ref{table:abla_study}. shows the results of our methods in comparison with the above five baselines by evaluating on ILSVRC dataset. Our method outperforms all five baselines by mAP and PDQ metrics. At frame level, our methods achieves frame level mAP of $72.1$ and video level mAP of $75.3$ by a margin of $0.2-0.5$ over the second best methods. In the measurement of PDQ, our methods achieve frame level of $39.4$ and video level of $40.2$ by a margin of $0.2-0.4$ over the second base method. Our method has a large margin of performance gain around $8.1$ over the baseline of Single R-FCN by PDQ metrics. In the baseline of Frame Bayesian, the performance could outperform Single R-FCN by a margin of $3.7$ in PDQ by naive inference on a frame-level prior. This performance gain suggests that Greedy NMS is detrimental to the discriminative power of R-FCN. Because it discards a wide spectrum of information that is helpful on distinguishing positive/negative bounding box and uncertainties of object's bounding box and categorical states. Greedy D$\&$T has nearly the same and even a worse of $1.6$ performance in mAP by comparing with Single R-FCN methods. This a little worsened performance may due to the fact of incorrect fetching between convolution feature maps and correlation kernel. Kalman-Link method reaches the performance only second to our proposed methods for its ability to infer on the uncertainties of locations from the states of its previous trajectories. However its previous trajectories is established by greedy forward matching of object cross frames in a deterministic way. By considering the object matching uncertainties under a uniform prior assumptions on object appearance and associations, our performance could achieve a $1.6$ gain in PDQ from Kalman-Link. Actually, as we consider the uncertainties in object linking and its states jointly, our algorithm allows for inferring object linking uncertainties reversely by particle reweighting.\\
\begin{table}
\caption{Comparison of our methods with 5 different kinds of baseline. Each baseline removes one or some of modules in our Bayesian framework and replaces with a naive non-Bayesian one. Our Bayesian one outperforms the baseline methods in all categories by introducing uncertainty treatment.}
\begin{center}
\begin{tabular}{|l|c c | c c |}
\hline
 &     & \hspace*{-3em}Frame  &   & \hspace*{-3em}Video
	    \\ \hline 
		  & \hspace*{-0.5em}  PDQ  & \hspace*{-0.5em}  mAP &\hspace*{-0.5em}   PDQ &\hspace*{-0.5em}  mAP
		 \\ \hline
		Our Methods  & \bf{39.4}  & \bf{72.1}  & \bf{40.2}  & \bf{75.3}
		\\ \hline
		Single R-FCN  & 31.7  & 70.3   & 32.1  & 70.9
		\\ \hline
		Frame Bayesian &  35.4  & 71.2   & 35.9  & 72.1
		\\ \hline 
		Greedy R-FCN  &  31.7  & 70.3   & 32.4  & 72.3
		\\ \hline
		Greedy D$\&$T & 31.8  & 68.7   & 32.4  & 72.7
		\\ \hline
		Kalman-Link & 37.8 & 71.9  &  39.4 & 74.9
      \\ \hline
\end{tabular}
\end{center}
\label{table:abla_study}
\end{table}
\vspace{-1em}
\subsection{Semi-Supervised Detection Result}
 We apply our methods to semi-supervised learning in M2Cai16-Tool-Location Dataset. In our implementation, we take another 3 consecutive frames from the labeling frames in a random order (forward and backward). We train on supervised learning loss in the first stage of our training epochs. And we add our semi-supervised loss term after our supervised training loss converges. Table \ref{table:semi}. shows our semi-supervised detection result in comparison with the one including supervised term only. Our semi-supervised learning algorithm achieves minor improvements over supervised learning from labeled frames only. Our method observes more obvious improvements on objects with low mAP.
 \begin{table}
	\centering
	\caption{Comparison of our semi-supervised learning algorithm with learning on supervised frames only.}
	\begin{tabular}{| l | c  c | c  c| }
	\hline
	    &     & \hspace*{-3em}Supervised  &   & \hspace*{-3em}Semi-Supervised
	    \\ \hline 
		  & \hspace*{-0.5em}  PDQ  & \hspace*{-0.5em}  mAP &\hspace*{-0.5em}   PDQ &\hspace*{-0.5em}  mAP
		 \\ \hline
		Grasper  & 35.4 & 46.2  & 37.4 & 52.3
		\\ \hline
		Bipolar  & 51.3    & 65.9   & 54.2  &  67.1
		\\ \hline
		Hook & 63.9    &  78.4  & 64.1  & 78.6
		\\ \hline 
      Scissors	 &  50.2  & 66.8   & 54.3  &  69.1
		\\ \hline
		Clipper  & 69.8 & 85.4  & 70.2 & 85.5
		\\ \hline
		Irrigator  & 11.3  & 16.2   & 14.9  & 23.5
		\\ \hline
		Specimen Bag  & 60.7  & 75.8  & 63.1  &   76.2
		\\ \hline	
	\end{tabular}
	\label{table:semi}
\end{table}
 
\vspace{-1em}
\section{Conclusion}
In this paper, we present our Bayesian model for multi-object detection and tracking in videos. Our method has shown the potential of formulating neural network model in a probabilistic way, especially for tasks that need to infer under uncertainties. 

{\small
\bibliographystyle{ieee_fullname}
\bibliography{egbib}
}

\end{document}